\documentclass[runningheads]{llncs}

\usepackage{eccv}

\usepackage{eccvabbrv}

\usepackage{graphicx}
\usepackage{booktabs}

\usepackage[accsupp]{axessibility}  

\usepackage{bbding} 
\usepackage{multirow} 
\usepackage{pifont}
\usepackage{bm} 
\usepackage{multirow}
\usepackage{colortbl}
\usepackage{colortbl}
\usepackage{color}
\usepackage{url}
\usepackage{bbding}
\usepackage{bm} 
\newcommand{\myparagraph}[1]{\textbf{#1}}
\definecolor{aliceblue}{rgb}{0.87, 0.92, 0.96}
\newcommand{\largemodel}[1]{\color{gray}{#1}}
\usepackage{stfloats} 
\definecolor{upcolor}{rgb}{0.73, 0.15, 0.12}
\definecolor{downcolor}{rgb}{0.02,0.24,0.74}
\usepackage{fontawesome5}
\usepackage{amsmath}
\usepackage{algorithm}
\usepackage{algpseudocode} 

\usepackage{caption}     
\usepackage{subcaption}  
\usepackage{graphicx}
\usepackage{wrapfig}

\usepackage{hyperref}
\definecolor{urlpink}{HTML}{FF4FA3}

\usepackage{orcidlink}

\begin{document}

\title{MediRound: Multi-Round Entity-Level Reasoning Segmentation in Medical Images} 

\titlerunning{Multi-Round Medical Reasoning Segmentation}

\author{Qinyue Tong\inst{1}\orcidlink{0009-0003-1007-9349} \and
Ziqian Lu\inst{2}\orcidlink{0009-0007-3579-9130} \and
Jun Liu\inst{1}\orcidlink{0009-0007-4759-2276} \and
Rui Zuo\inst{1}\orcidlink{0009-0003-9982-9221} \and
Zhe-Ming Lu\inst{1(}\Envelope\inst{)}\orcidlink{0000-0003-1785-7847} \and
Yueming Jin\inst{3(}\Envelope\inst{)}\orcidlink{0000-0003-3775-3877}}

\authorrunning{Q.~Tong et al.}

\institute{Zhejiang University, Hangzhou, Zhejiang, China \\
\email{\{qinyuetong,junliu0930,ruizuo,zheminglu\}@zju.edu.cn}
\and Zhejiang Sci-Tech University, Hangzhou, Zhejiang, China\\ \email{ziqianlu@zstu.edu.cn} \and National University of Singapore, Singapore\\ \email{ymjin@nus.edu.sg}
}

\maketitle

\begin{center}
    \captionsetup{type=figure}
    \includegraphics[width=1\linewidth]{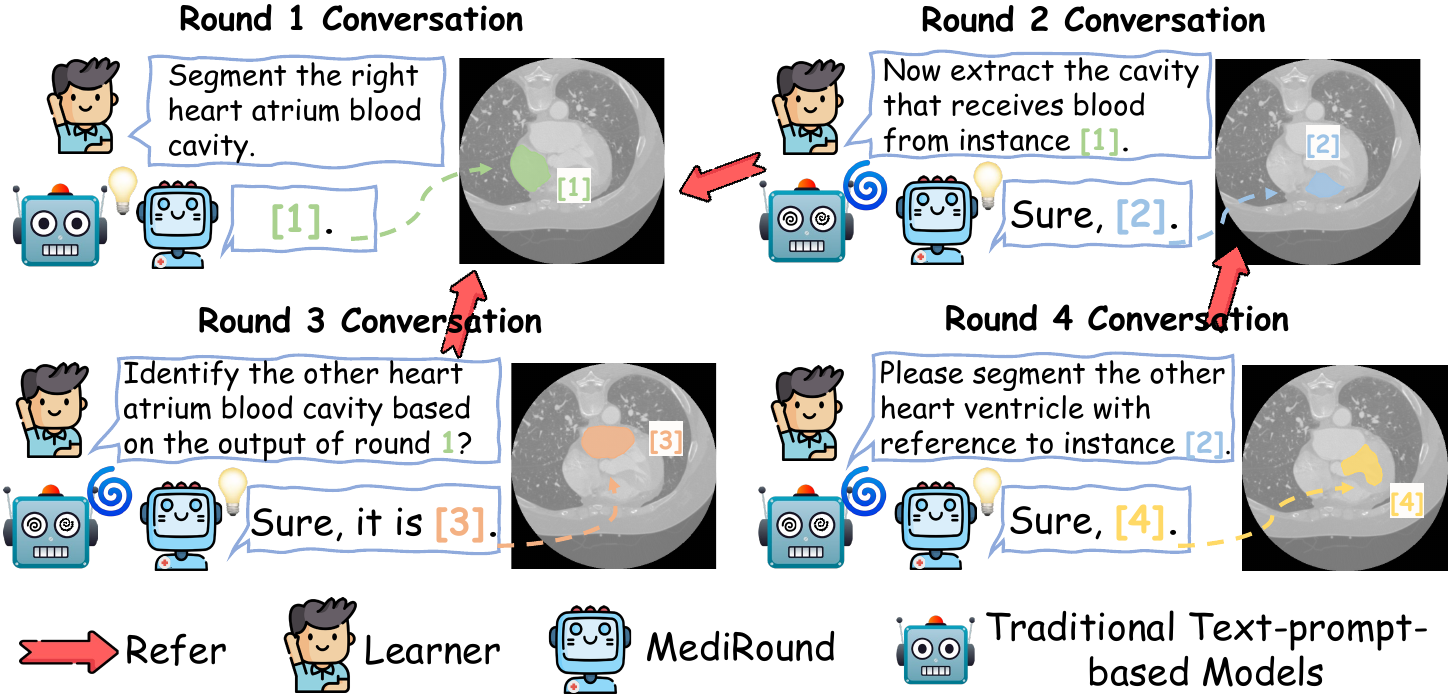}
    \captionof{figure}{
    A medical education dialogue illustrating the proposed MediRound. Our model can comprehend user queries that refer to the mask results from previous rounds (\textit{e.g., the Round 2 query refers to the Round 1 mask result}), enabling cross-round entity-level reasoning in multi-round medical conversations. 
    This iterative querying paradigm allows learners to progressively develop their understanding of medical entities.
In contrast, conventional text-prompt-based medical segmentation methods struggle with such a complex and context-dependent task.
    }
    \label{fig:teaser}
\end{center}

\begin{abstract}
Despite notable progress in text-guided medical image segmentation nowadays, these methods are limited to single-round dialogues and fail to support multi-round reasoning, which is important for \textbf{medical education} scenarios.
In this work, we introduce Multi-Round Entity-Level Medical Reasoning Segmentation (\textbf{MEMR-Seg}), a new task that requires generating segmentation masks through multi-round queries with entity-level reasoning, helping learners progressively develop their understanding of medical knowledge.
To support this task, we construct \textbf{MR-MedSeg}, a large-scale dataset of 177K multi-round medical segmentation dialogues, featuring entity-based reasoning across rounds.
Furthermore, we propose \textbf{MediRound}, an effective baseline model designed for multi-round medical reasoning segmentation.
To mitigate the inherent error propagation within the chain-like pipeline of multi-round segmentation, we introduce a lightweight yet effective \textbf{Judgment \& Correction Mechanism} during model inference.
Experimental results demonstrate that our method effectively addresses the MEMR-Seg task and outperforms conventional medical referring segmentation methods.
\textit{The project is available at 
\href{https://github.com/Edisonhimself/MediRound}
{\textcolor{urlpink}{\nolinkurl{https://github.com/Edisonhimself/MediRound}}}.}

\end{abstract}

\section{Introduction}
\label{sec:intro}
Medical image segmentation focuses on identifying and outlining key anatomical or lesion-related regions across diverse medical imaging modalities~\cite{instruction-med-seg-onesentence-1,intro_first_seg-2,intro_first_seg-3}.
This capability underpins a broad range of downstream medical applications, including clinical analysis, medical education, and biomedical research ~\cite{intro-disease-progression-tracking-1,intro-disease-progression-tracking-2,intro-disease-progression-tracking-3,intro-disease-progression-tracking-4, intro-medical-researches-1}.
Most existing studies focus predominantly on task-specific segmentation~\cite{intro-specialist-model-1,intro-specialist-model-2,related-works-unet}, commonly referred to as ``specialist models''.
Despite their remarkable performance, these models generally suffer from limited adaptability and lack the interactive capabilities necessary for practical, real-world usage.

To address this issue, recent text-prompt-based medical image segmentation methods \cite{biomed-parse, imis_model, huang2024_ReclMIS, li2023lvit} enable users to specify target regions through natural language instructions, allowing segmentation to be guided more flexibly according to user intent.
Building upon these developments, several MLLM-based approaches \cite{tong2025medisee, yan2025medreasoner, huang2025medsegR} further enhance the interactivity and practicality by enabling reasoning-driven medical segmentation, allowing users to obtain the desired masks through implicit queries.

However, these models primarily focus on single-round medical segmentation and are incapable of supporting multi-round and continuous text-based interactions.
Although the one-shot querying paradigm can be efficient in clinical use, it presumes high-quality prompts and substantial user expertise. As a result, it is well suited to medical experts but less conducive to the learning process of \textbf{non-expert users}. In \textbf{medical education} settings, visual understanding is typically developed \textbf{progressively}.
As illustrated in Figure \ref{fig:teaser}, learners tend to formulate new segmentation queries based on the mask results obtained from previous rounds. 
For example, in \textit{Round 2 Conversation}, the user’s query is based on the mask generated in \textit{Round 1 Conversation}, and the results of \textit{Round 2 Conversation} in turn serve as a reference for \textit{Round 4 Conversation}.
Such iterative questioning helps learners progressively develop a better visual understanding of different medical entities and grasp the relationships among them more effectively.
Moreover, these follow-up queries are often relation-driven and logically constrained, reflecting how trainees learn anatomy and pathology through relational concepts rather than precise standalone descriptions, which aligns with how learning occurs in real-world medical education.
Nevertheless, due to the lack of multi-round entity-level reasoning capabilities, existing methods often fail to produce the expected responses when dealing with inputs that involve cross-round logical dependencies.

In this work, we define a new medical vision-language task, termed \textbf{MEMR-Seg} (Multi-Round Entity-Level Medical Reasoning Segmentation), which entails generating binary segmentation masks based on multi-round medical image queries, involving entity-level reasoning across dialogue rounds.
Notably, the queries are not only multi-round in nature, but each round is also a derivative and extended inquiry based on the entity results from the previous round.
To successfully perform this new task, the model should possess two crucial abilities: 
\textit{(1) supporting multi-round dialogue and cross-round entity-level reasoning}; and 
\textit{(2) generating accurate segmentation results in response to user queries}.

As data scarcity remains a critical challenge in accomplishing MEMR-Seg, we first propose a Multi-Round Entity-Level Medical Reasoning Segmentation dataset, termed \textbf{MR-MedSeg}, which contains a large number of medical conversations centered on entity-level reasoning segmentation with multi-round queries and answers.
We collect all metadata from the publicly accessible SA-Med2D-20M \cite{sam_med_2d_20m}. Subsequently, inspired by SegLLM \cite{multi-round-seg}, we construct multi-round question–answer pairs through a combination of manual annotation and GPT-5-based generation.
Consequently, the MR-MedSeg dataset comprises a large and diverse collection of 177K conversations.
It is worth noting that a major difference between our MR-MedSeg and traditional medical segmentation datasets \cite{intro-traditional-seg-dataset-1,intro-traditional-seg-dataset-2,intro-traditional-seg-dataset-3,traditional-seg-dataset-1,traditional-seg-dataset-2,tong2025medisee} lies in the fact that our dataset not only features multi-round dialogues but also incorporates entity-based reasoning logic across different dialogue rounds.
This design enhances the educational utility of MR-MedSeg and facilitates more interactive medical image segmentation.

Furthermore, building upon existing reasoning-based segmentation works \cite{intro-seg-llm,lisa,lisa++, multi-round-seg}, we introduce \textbf{MediRound}, an effective baseline model designed for multi-round medical reasoning segmentation.
MediRound integrates the segmentation feature representations from the reference round and the textual information from previous dialogues with the current query. 
These are jointly embedded into the input sequence of the inner-MLLM.
This design enables the model not only to comprehend the current query but also to access the core information from the reference round while maintaining awareness of the entire conversational history.
Moreover, to address the inevitable issue of error accumulation inherent in multi-round segmentation (\textit{as illustrated in Figure \ref{fig:teaser}, if the result of Round 1 contains errors, these errors may propagate to Round 2, and the resulting inaccuracies in Round 2 can further affect Round 4}), we introduce a lightweight yet effective \textbf{Judgment \& Correction Mechanism} during the model inference stage. 
Applied after the end-to-end model training phase, this mechanism helps MediRound mitigate error propagation in practical multi-round inference and effectively enhances the overall accuracy of multi-turn entity-level reasoning.
Extensive experimental results demonstrate the superior performance of MediRound in both multi-round and single-round medical segmentation, while confirming the effectiveness of the Judgment \& Correction Mechanism.
Overall, we summarize our main contributions as follows:
\begin{itemize}
    \item We introduce the Multi-Round Entity-Level Medical Reasoning Segmentation (\textbf{MEMR-Seg}) task, which requires reasoning over multi-round queries that refer to results from earlier rounds in medical images.
    \item We construct the \textbf{MR-MedSeg} dataset, which comprises 177K multi-round medical segmentation dialogues that require multi-round reasoning. 
    \item We introduce an effective baseline model, \textbf{MediRound}, and propose a \textbf{Judgment \& Correction Mechanism} to address the inherent error accumulation during multi-round reasoning segmentation. 
    Comprehensive experiments verify the effectiveness of our model and the utility of the proposed mechanism.
\end{itemize}

\section{Related Works}
\noindent \myparagraph{Medical Image Segmentation.}
As one of the foundational architectures in medical image segmentation, the U-Net family \cite{related-works-unet,related-works-unet++,related-works-ResU-Net,related-works-ResU-nnU-Net} has inspired numerous extensions and variants. 
Despite their strong performance on specific tasks and imaging modalities, these fully automated and task-specific models often lack the flexibility and user interactivity necessary for adaptability in practical, real-world medical environments.
Text-prompt-based segmentation models address this by allowing users to guide segmentation via language prompts \cite{segvol,related-works-text-1,imis_model,biomed-parse}, and recent works further support complex implicit queries \cite{tong2025medisee,huang2025medsegR,yan2025medreasoner}, improving flexibility and interactivity. 
However, they are unable to handle continuous, multi-turn queries or perform cross-round reasoning over medical entities, both of which are common and crucial in medical education scenarios.
This work primarily centers on multi-round entity-based medical reasoning segmentation, enabling entity-centered reasoning across dialogue rounds while generating accurate segmentation masks.

\noindent \myparagraph{Reasoning Segmentation.}
General reasoning segmentation is introduced by LISA \cite{lisa}, which interprets implicit textual prompts to generate segmentation masks by combining MLLMs’ vision-language understanding \cite{llava} with SAM’s segmentation capabilities \cite{segment-anything}.
More recently, several follow-up studies extend reasoning segmentation \cite{mmr-iclr2025,glamm-cvpr2024,perceptiongpt-cvpr2024,next-chat}. Among them, SegLLM \cite{multi-round-seg} represents the initial effort to address multi-round reasoning segmentation in natural images.
Compared to natural image domains, multi-round reasoning segmentation in medical imaging could provide enhanced practical utility. 
This is because such a segmentation paradigm aligns well with the process through which trainees learn medical knowledge.
However, despite recent progress in medical reasoning segmentation \cite{tong2025medisee,huang2025medsegR,yan2025medreasoner,xing2026medvl3d}, this aspect remains unaddressed.
Concurrent with our work, MTurn-Seg \cite{MTurn-Seg} introduces a bilingual benchmark for multi-turn medical image dialogue that incorporates segmentation.
In this work, building upon prior works \cite{llava-med,medsam_model,sam_med_2d_20m,multi-round-seg}, we develop a method for multi-round reasoning in medical image segmentation.

\begin{figure}[t]
    \centering
    \includegraphics[width=1\linewidth]{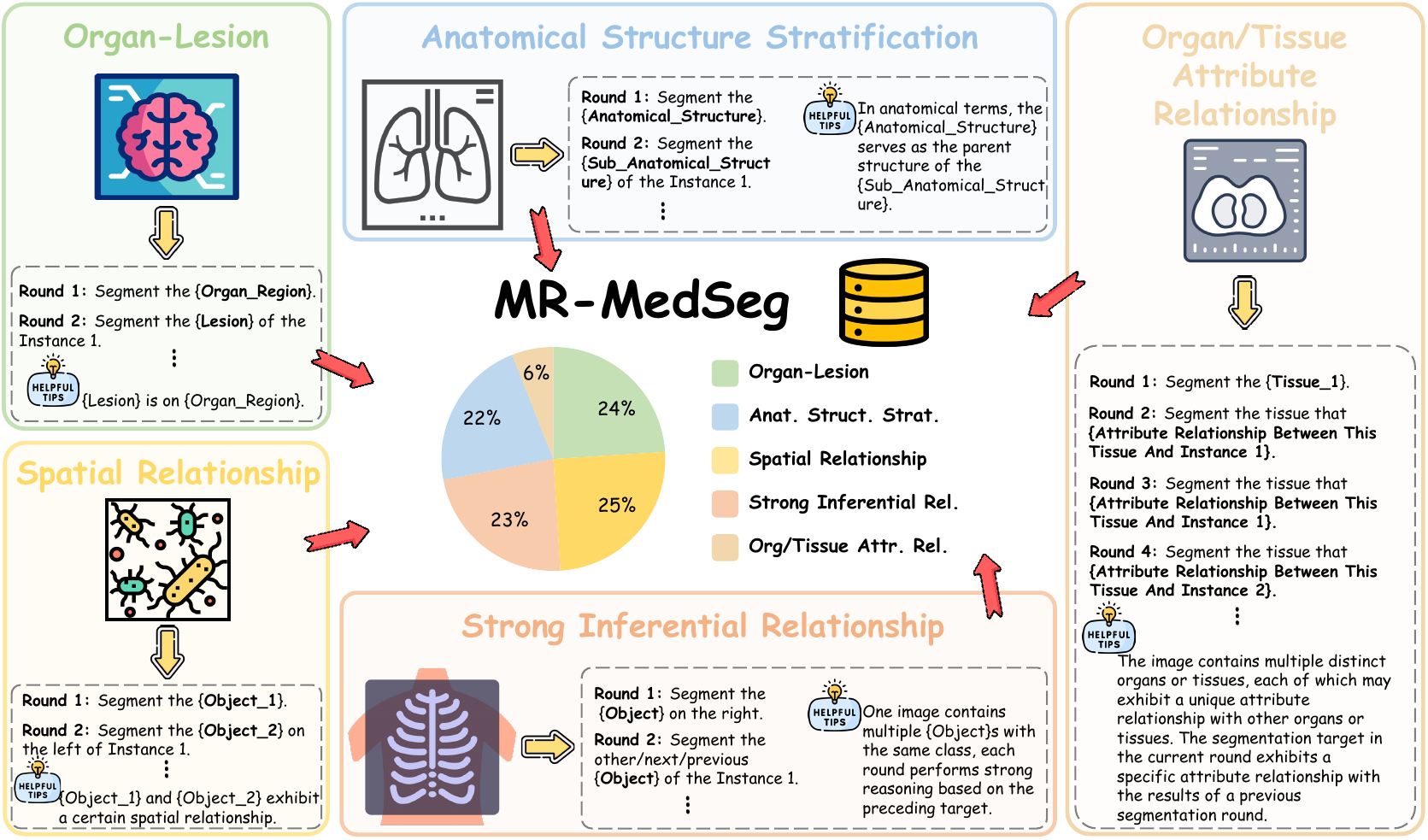}
    \caption{
Overview of MR-MedSeg. Our dataset comprises five types of medical reasoning dialogues, each characterized by a specific form of inter-instance relationship, encompassing nearly all multi-round segmentation scenarios in medical education settings.
    }
    \label{fig:dataset_overview}
\end{figure}

\begin{figure*}[t]
    \centering
    \includegraphics[width=0.85\linewidth]{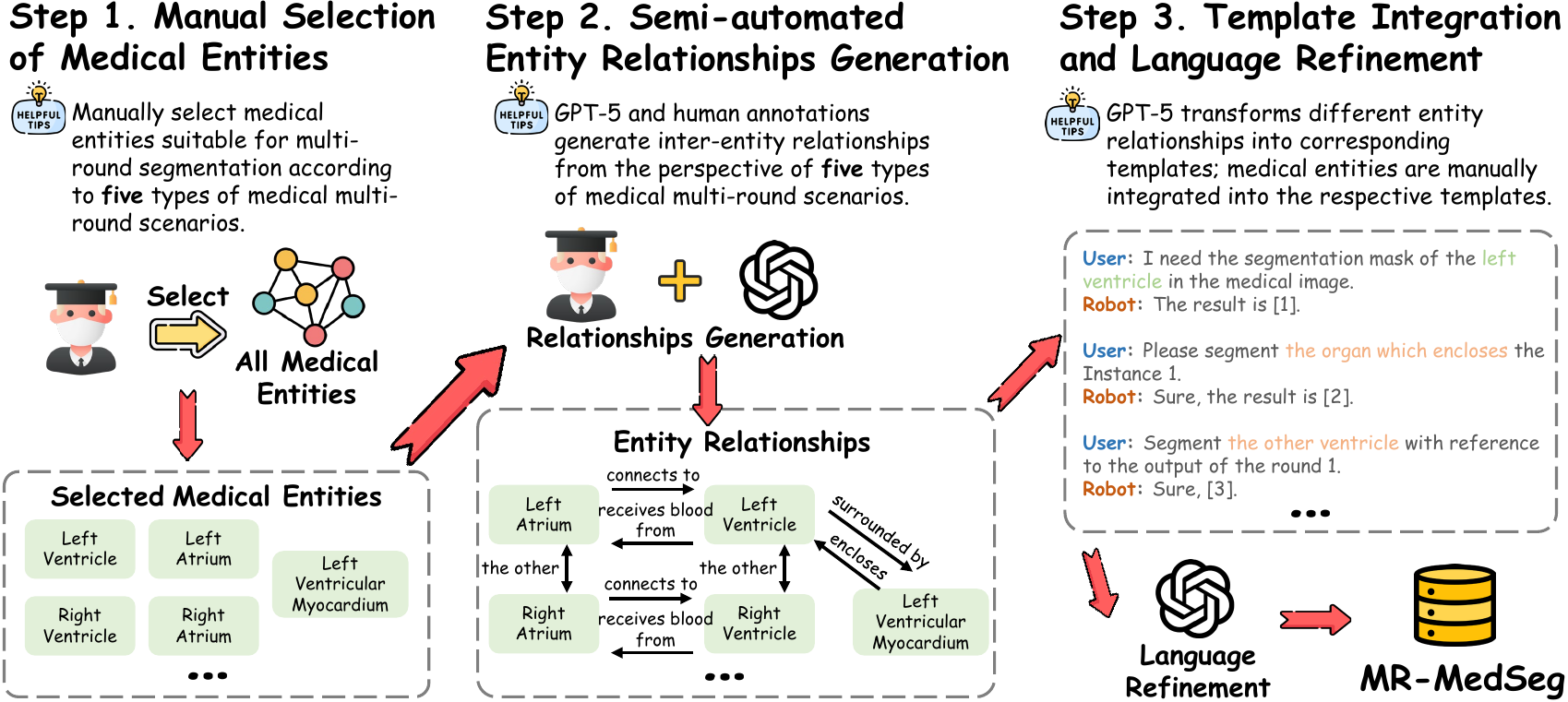}
    \caption{
    Semi-automatic pipeline for constructing MR-MedSeg dataset. The process includes three stages: entity selection, relationship generation, and template integration. 
    We construct the pipeline with manual annotation, augmented by GPT-5 generation.
    }
    \label{fig:data_pipeline}
\end{figure*}
\section{MR-MedSeg}

\noindent \myparagraph{Dataset Overview.}
The overview of our proposed MR-MedSeg dataset is shown in Figure~\ref{fig:dataset_overview}.
As illustrated in the figure, we design five representative types of interaction scenarios tailored for multi-round entity-level medical reasoning segmentation, each reflecting realistic use cases in medical education practice.
These scenarios each involve distinct forms of inter-instance relationships, including organ–lesion dependency, anatomical hierarchy, organ/tissue attribute association, spatial relationship, and strong inferential relationship.
Figure~\ref{fig:dataset_overview} further presents the overall composition and proportions of these scenarios in MR-MedSeg.
\noindent \myparagraph{Data Construction Pipeline.} 
As shown in Figure \ref{fig:data_pipeline}, we design a semi-automatic pipeline for constructing the MR-MedSeg dataset. 

In \textit{Step 1}, we manually select appropriate medical entities according to the five types of multi-round medical interaction scenarios illustrated in Figure \ref{fig:dataset_overview}. Specifically, we traverse the SA-Med2D-20M dataset \cite{sam_med_2d_20m} and identify sub-datasets that meet the criteria of each scenario. From these selected sub-datasets, we further manually select the medical entities that best match the five designed interaction settings. 
MR-MedSeg is built upon the curated image, mask, and label annotations of SA-Med2D-20M, which substantially alleviates the burden of data cleaning, preprocessing, and manual annotation.
Next, in \textit{Step 2}, we construct diverse inter-entity relationships for the selected medical entities based on the entity relationships defined across five scenarios, employing a hybrid strategy that combines \textit{manual annotation} with \textit{GPT-5-based generation}.
Notably, when generating relationships that are independent of image information, such as biological or anatomical attributes shared among medical entities which are generally consistent across images, we omit the reference to the image context. In contrast, when generating relationships that depend on image information, such as the relative spatial positions of anatomical structures that vary between images, we explicitly incorporate the corresponding image context to ensure both accuracy and contextual relevance.
Subsequently, in \textit{Step 3}, we employ GPT-5 \cite{gpt-5} to transform each entity relationship into 50–80 diverse yet semantically equivalent templates. 
The corresponding medical entities are then manually integrated into these templates to construct the initial version of the multi-round medical segmentation data. 
Finally, we perform an additional automatic refinement process using GPT-5 to further enhance the fluency and accuracy of the generated dialogues.
Moreover, we engage three inspectors with professional medical expertise to screen and filter the data in MR-MedSeg, thereby maximizing the validity and reliability of the dataset.

\noindent \myparagraph{Data Statistics.} 
The MR-MedSeg dataset comprises over 177K multi-round medical segmentation dialogues that require cross-round reasoning, constructed from an initial candidate pool of 118K images and 569K masks.
We assign 174,934 conversations to the training set, 1,270 to the validation set, and 1,273 to the test set.
MR-MedSeg further covers 168 distinct medical entities, spanning a wide range of anatomical structures, pathological findings, and tissue types.
The dataset also covers 9 medical imaging modalities, further highlighting its diversity. 

\section{MediRound}
\label{sec:MediRound}
We propose an effective baseline for multi-round medical reasoning segmentation, termed MediRound.
MediRound demonstrates the ability to perform multi-round reasoning over segmented medical entities by understanding information from preceding rounds to achieve fine-grained segmentation in the current round.
To address the inherent drawback that errors in previous rounds can severely affect the performance of subsequent rounds in the multi-round segmentation pipeline, we introduce a novel Judgment \& Correction Mechanism (JCM).
Details of MediRound and JCM follow.

\subsection{Model Architecture}
\begin{figure*}[t]
    \centering
    \includegraphics[width=0.85\linewidth]{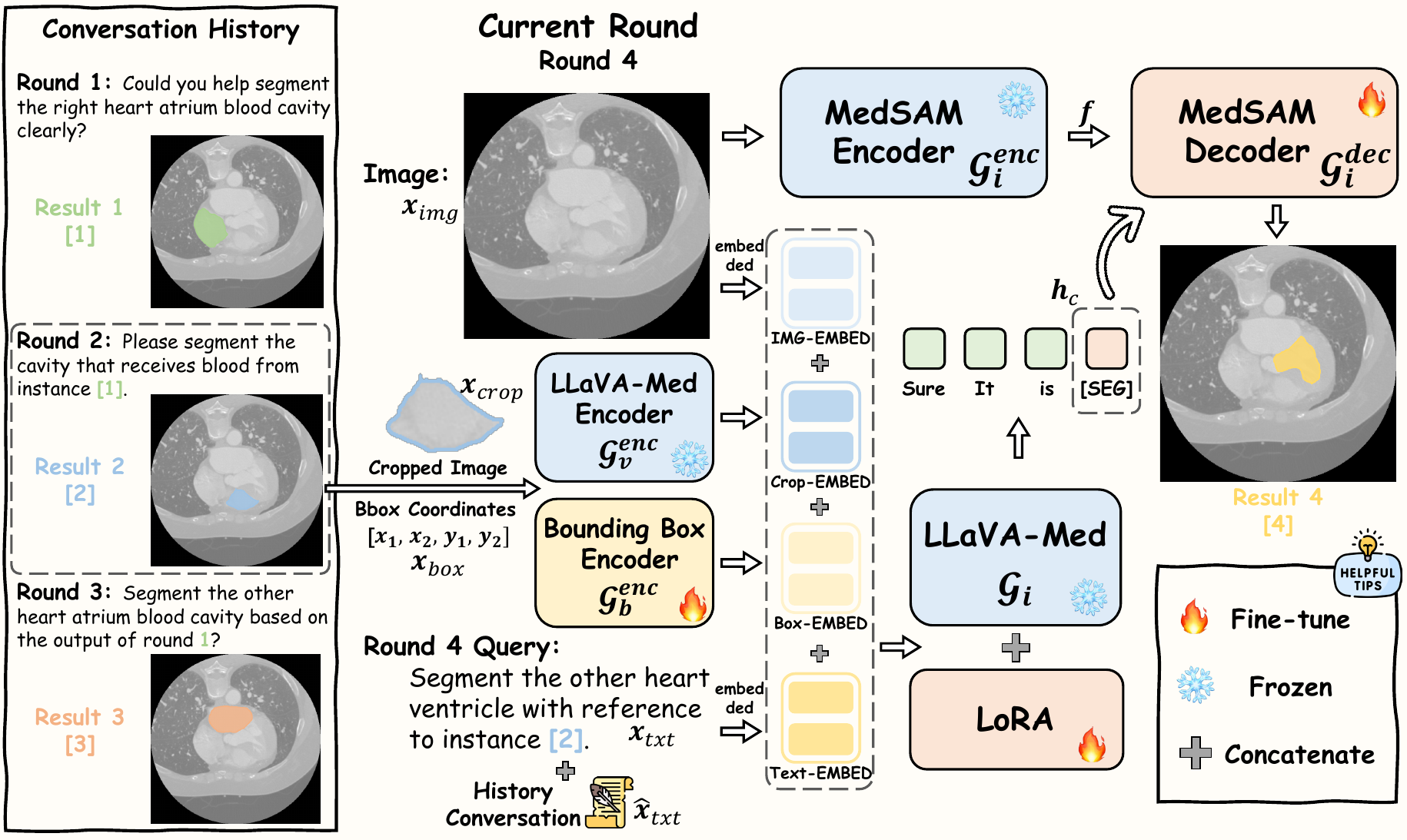}
    \caption{
Overview of the MediRound framework. The figure illustrates the model’s workflow in processing the Round 4 conversation, referring to the Round 2 mask. The encoder $\mathcal{G}_v^{enc}$ consists of the vision encoder and visual projection layer from LLaVA-Med.
}
    \label{fig:model_overview}
\end{figure*}

Figure~\ref{fig:model_overview} illustrates the model architecture of our proposed MediRound, which is composed of the following modules: MedSAM~\cite{medsam_model} serving as the vision backbone with an image encoder $\mathcal{G}_i^{enc}$ and a mask decoder $\mathcal{G}_i^{dec}$; LLaVA-Med~\cite{llava-med} functioning as the multimodal large language foundation model $\mathcal{G}_i$; 
the LLaVA-Med visual encoder $\mathcal{G}_v^{enc}$ for extracting and projecting image embeddings; 
and a linear projection acting as the bounding box encoder $\mathcal{G}_b^{enc}$.

\subsection{Multi-Round Reasoning and Segmentation}
Inspired by existing MLLM-based segmentation methods \cite{lisa++,mmr-iclr2025,next-chat} and the pioneer in general multi-round segmentation method SegLLM \cite{multi-round-seg}, we propose MediRound tailored for multi-round medical reasoning segmentation.
Specifically, we first extend the vanilla vocabulary of LLM by introducing a special token, [SEG], which indicates a request to produce the segmentation mask.
As shown in Figure~\ref{fig:model_overview}, \textbf{three rounds of conversations have already been established}. 
Given the \textbf{Round 4} query \(x_{txt}\) together with the original input image \(x_{img}\) and history conversation \(\hat{x}_{txt}\) (including the question–answer pairs from Rounds 1, 2, and 3), 
we first utilize the mask output from the \textbf{reference round (Round 2)} to crop the corresponding object \(x_{crop}\) from the original image. 
The cropped region and its bounding box coordinates \(x_{box}\) in the original image are used as reference information for the referred medical entity. 
Subsequently, the LLaVA-Med visual encoder \(\mathcal{G}_v^{enc}\) and the bounding box encoder \(\mathcal{G}_b^{enc}\) are employed to extract their respective features, which are then concatenated with \(x_{txt}\), \(\hat{x}_{txt}\) and \(x_{img}\) to form the complete input.
This input design enables the model not only to comprehend the current query but also to access key information from the reference round while maintaining awareness of the full conversational history.
We then feed the complete input into LLaVA-Med \(\mathcal{G}_i\), which in turn outputs a text answer \(\hat{y}_{txt}\). 
It can be formulated as (where $x_{img}$, $x_{txt}$, $\hat{x}_{txt}$ also denote their embeddings):
\begin{equation}
    \hat{y}_{txt} = \mathcal{G}_i(x_{img},  \mathcal{G}_v^{enc}(x_{crop}),  \mathcal{G}_b^{enc}(x_{box}), x_{txt}, \hat{x}_{txt}).
\end{equation}
The output \(\hat{y}_{txt}\) includes the special token [SEG] when the model attempts to segment a specific medical entity within \(x_{img}\).
We then extract the last-layer embedding of the LLM corresponding to the [SEG] token, denoted as \(h_{c}\), which encapsulates the rich semantic representation of the medical entity inferred by the MLLM. 
The vision backbone \(\mathcal{G}_i^{enc}\) first extracts dense visual representations \(f\) from the image \(x_{img}\). These features are then integrated with \(h_{c}\) by the decoder \(\mathcal{G}_i^{dec}\) to predict the target mask \(\hat{\mathbf{M}}\).
The process can be formulated as:
\begin{align}
\begin{aligned}
     f = \mathcal{G}_i^{enc}&\;(x_{img}), \quad               
    \hat{\mathbf{M}} = \mathcal{G}_i^{dec}(f, h_{c}).
\end{aligned}
\end{align}


\begin{algorithm}[t]
\caption{\small Multi-Round Inference of MediRound with JCM}
\label{alg:multi_round_seg_jcm}
\scriptsize
\begin{algorithmic}[1]
\Require Image $x_{img}$; multi-round queries $\{x_{txt}^{(t)}\}_{t=1}^T$; $\mathcal{G}_i$: LLaVA-Med; $\mathcal{G}_i^{dec}$: MedSAM decoder; $h_c$: [SEG] feature; $\text{MLP}_J$: Quality Judgment Module; $\text{MLP}_C$: Correction Module; $\beta$: the threshold used by Quality Judgment Module to judge the quality of [SEG] feature
\Ensure Predicted masks $\{\hat{\mathbf{M}}^{(t)}\}_{t=1}^T$, with initialization $\hat{\mathbf{M}}^{(0)} \gets \emptyset$
\For{$t = 1$ to $T$}
\State $r \gets \text{In this round, user selects the previous round } r \text{ mask from } \{0, 1, ..., t-1\}$ for reference; $r=0$ means no mask is referenced
    \State $\tilde{x}_{txt}^{(t)} \gets \text{IntegrateMaskInfo}(x_{txt}^{(t)}, \hat{\mathbf{M}}^{(r)})$
    \State $h_c \gets \mathcal{G}_i(x_{img}, \tilde{x}_{txt}^{(t)})$
    \State $q \gets \text{MLP}_J(h_c)$
    \If{$q > \beta$} 
        \State $\hat{\mathbf{M}}^{(t)} \gets \mathcal{G}_i^{dec}(h_c)$
    \Else
        \State $h_c' \gets \text{MLP}_C(h_c)$; $\hat{\mathbf{M}}^{(t)} \gets \mathcal{G}_i^{dec}(h_c')$
    \EndIf
\EndFor
\end{algorithmic}
\end{algorithm}

\subsection{Judgment \& Correction Mechanism}
During the training of MediRound, we adopt teacher forcing \cite{teacher_forcing} and directly use the ground-truth masks to obtain the referred information. 
While this strategy effectively optimizes the model parameters, it introduces a discrepancy between the training and testing scenarios: \textit{during multi-round evaluation, the trained model can only rely on its own outputs from previous rounds as reference information}.
As illustrated in Figure \ref{fig:model_overview}, the mask information from the referred round provides essential contextual guidance for the current round. 
\begin{wrapfigure}{r}{0.5\columnwidth}
    \centering
    \includegraphics[width=1\linewidth]{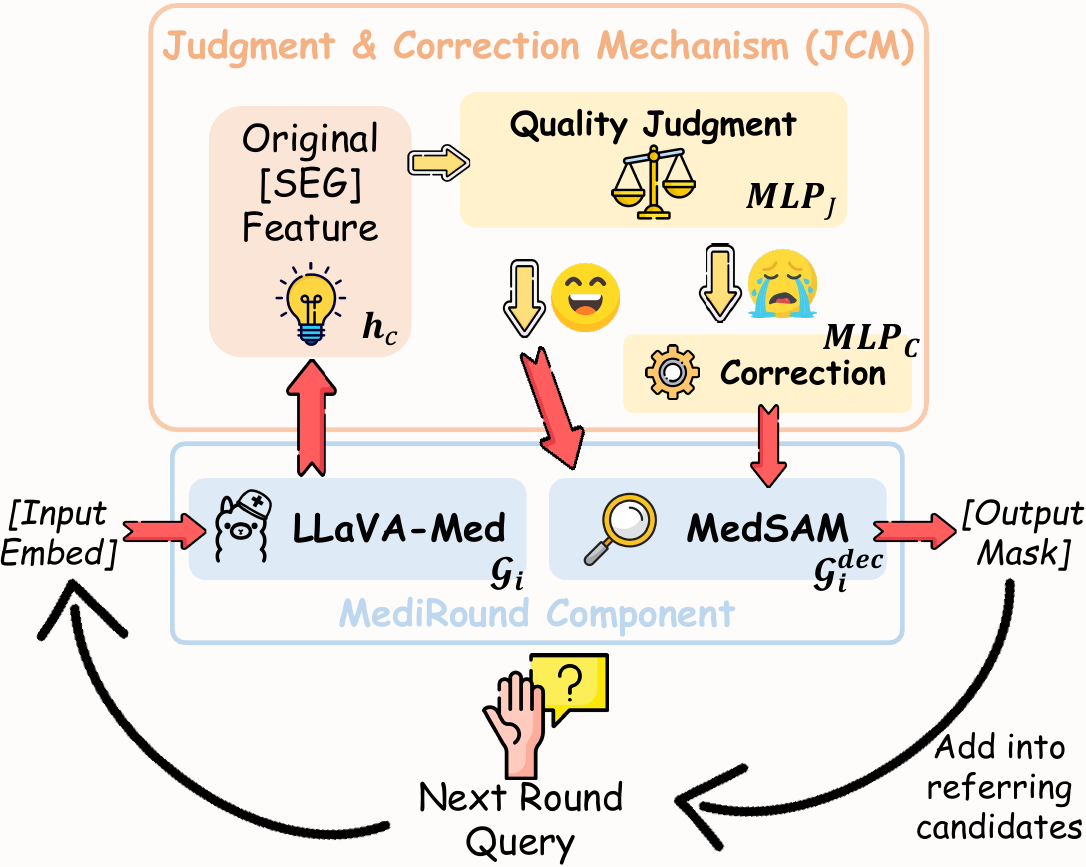}
    \caption{
     Illustration of how the Judgment \& Correction Mechanism assists MediRound in multi-round reasoning. This mechanism evaluates and optimizes the quality of the [SEG] hidden layer features in each round, effectively preventing current-round errors from being propagated to later dialogues. 
    }
    \label{fig:loop_pipeline}
\end{wrapfigure}
This dependency introduces a potential limitation: errors from earlier segmentation rounds may accumulate and propagate, ultimately degrading the accuracy of subsequent predictions during evaluation.
This effect may become increasingly pronounced as the number of segmentation rounds increases, due to the progressive accumulation and amplification of errors.
Therefore, we introduce a lightweight yet effective Judgment \& Correction Mechanism that enables the model to transfer higher-quality masks from previous rounds to subsequent ones during evaluation, thereby mitigating the accumulation of segmentation errors.
Motivated by prior studies \cite{lisa-READ,tong2025medisee} demonstrating that the [SEG] embedding encodes rich target semantics and that semantic drift within this embedding directly degrades mask quality, we intervene at the [SEG] embedding level to mitigate error propagation.
Specifically, as illustrated in Figure \ref{fig:loop_pipeline}, during the MediRound evaluation process, we employ a \textbf{Quality Judgment Module} to assess the quality of the hidden features associated with the output [SEG] token in the current round. 
If the feature quality is deemed unsatisfactory, the features are then passed to the \textbf{Correction Module} for feature refinement; otherwise, they are directly decoded into masks as reference candidates. 
In this way, when later rounds refer to the current round result, they receive a higher-quality reference mask, which facilitates more coherent cross-round reasoning.
To better illustrate how JCM assists MediRound, Algorithm~\ref{alg:multi_round_seg_jcm} provides the pseudocode. Notably, both the Quality Judgment and Correction Modules are lightweight MLPs, with training details in Sec.~\ref{sec:training_optimization}.

\subsection{Training and Optimization}
\label{sec:training_optimization}
\noindent \myparagraph{Training the MediRound.} 
In end-to-end training for MediRound, we use LoRA to conduct effective fine-tuning on the LLM in \(\mathcal{G}_i\), while freezing the vision backbone \( \mathcal{G}_i^{enc}\) and LLaVA-Med vision encoder \(\mathcal{G}_v^{enc}\).
The mask decoder \( \mathcal{G}_i^{dec}\) and the bounding box encoder \(\mathcal{G}_b^{enc}\) are fully trainable.

For optimization, MediRound is trained through two supervisions.
For the text generation objective, we optimize the Auto-Regressive Cross-Entropy loss \(\mathcal{L}_{txt}\) between the predicted textual response \(\hat{y}_{txt}\) and the ground-truth text answer \(y_{txt}\).
For the mask generation, the mask loss \(\mathcal{L}_{mask}\) is calculated between the predicted mask \( \hat{\mathbf{M}}\) and the ground-truth mask \( \mathbf{M}\). 
The mask loss \(\mathcal{L}_{mask}\) is formulated as a weighted combination of the per-pixel Binary Cross-Entropy \(\mathcal{L}_{bce}\) and the DICE loss \(\mathcal{L}_{dice}\), controlled by the weighting factors \(\lambda_{bce}\) and \(\lambda_{dice}\).
The overall loss \(\mathcal{L}_{MediRound}\) is formulated as:
\begin{equation}
\label{eq:mediround_loss}
\begin{split}
    &\mathcal{L}_{MediRound} = \mathcal{L}_{txt} + \mathcal{L}_{mask}, \\
    &\mathcal{L}_{mask} = \lambda_{bce}\mathcal{L}_{bce} + \lambda_{dice}\mathcal{L}_{dice}. \\
\end{split}
\end{equation}

\begin{figure*}[t]
    \centering
    \includegraphics[width=0.95\textwidth]{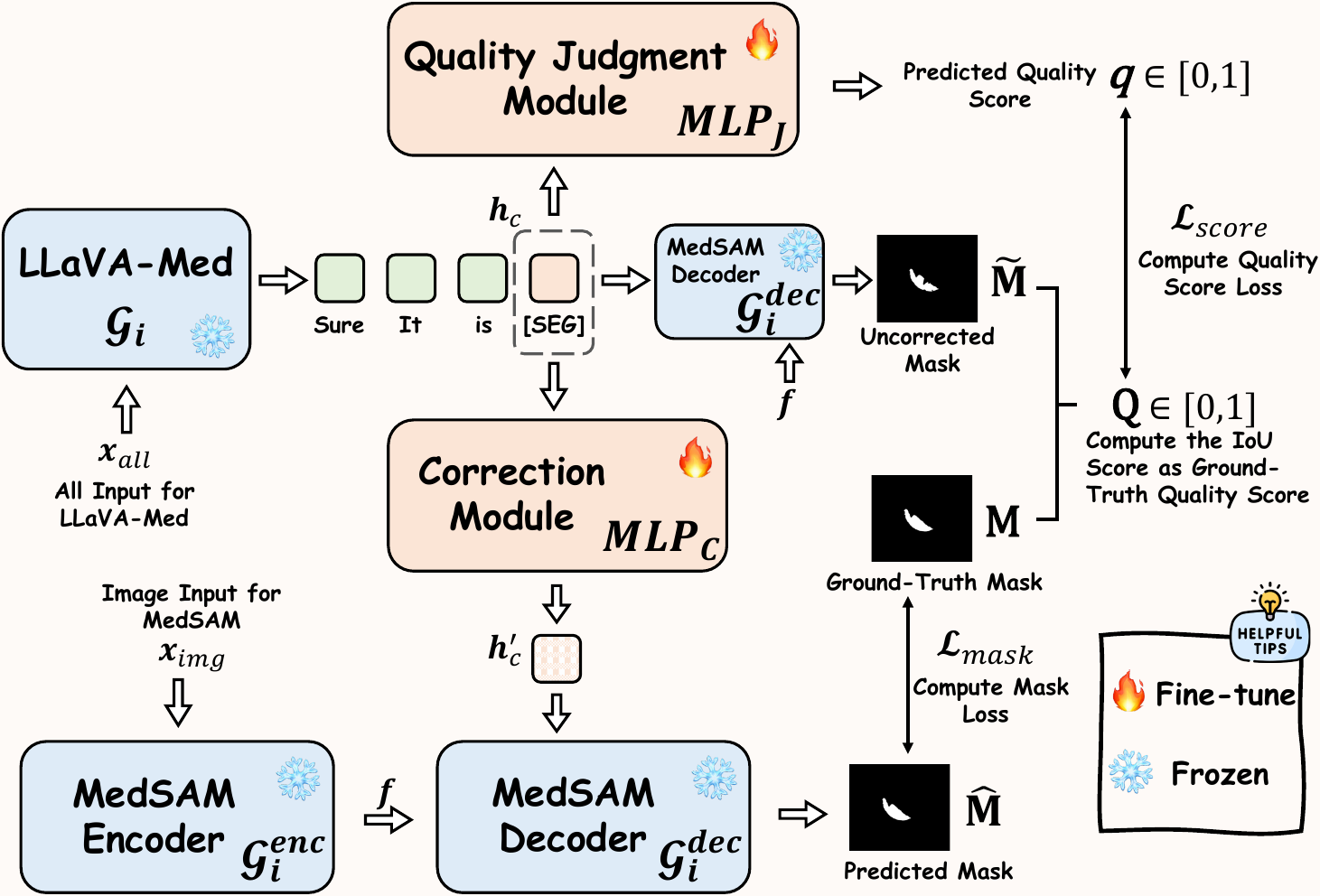}
    \caption{
     Overview of the JCM training pipeline.
    }
    \label{appendix_fig:jcm_train_pipeline}
\end{figure*}

\noindent \myparagraph{Training the JCM.} 
The overall training pipeline of JCM is illustrated in Figure \ref{appendix_fig:jcm_train_pipeline}.
As shown in Figure \ref{appendix_fig:jcm_train_pipeline}, LLaVA-Med takes $x_{all}$ (denoting the full set of inputs) and produces the [SEG] token as output. 
The hidden feature of [SEG], $h_{c}$, is first fed into the Correction Module $MLP_{C}$, producing $h'_{c}$.  
The mask decoder $\mathcal{G}_i^{dec}$ then takes $h'_{c}$ along with the visual features $f$ (extracted by the vision backbone $\mathcal{G}_i^{enc}$) and outputs the predicted mask $\hat{\mathbf{M}}$. 
In parallel, $\mathcal{G}_i^{dec}$ takes $h_c$, $f$ and outputs the uncorrected mask $\tilde{\mathbf{M}}$.
The process can be formulated as:
\begin{equation}
    h'_{c} = {MLP}_{C}(h_{c}), \quad \hat{\mathbf{M}} = \mathcal{G}_i^{dec}(f, h'_{c}), \quad \tilde{\mathbf{M}} = \mathcal{G}_i^{dec}(f, h_{c}).
\end{equation}
Additionally, we also feed $h_c$ into the Quality Judgment Module $MLP_J$, which outputs a quality score $q \in [0,1]$ for $h_c$:
\begin{equation}
q = MLP_J(h_c), \quad q \in [0,1]
\end{equation}

We utilize the end-to-end trained MediRound weights to optimize the Correction Module $MLP_C$ and the Quality Judgment Module $MLP_J$ within the JCM.  
During the training of JCM, all parameters of the MediRound modules \textit{(e.g., LLaVA-Med, MedSAM, etc.}) are frozen, and only the Correction Module $MLP_C$ and the Quality Judgment Module $MLP_J$ are trainable.

For optimization, as illustrated in Figure~\ref{appendix_fig:jcm_train_pipeline}, the Correction Module $MLP_C$ and the Quality Judgment Module $MLP_J$ are jointly optimized using the mask loss $\mathcal{L}_{mask}$ and the quality score loss $\mathcal{L}_{score}$.
For the quality score loss $\mathcal{L}_{score}$, we compute the Binary Cross-Entropy loss $\mathcal{L}_{bce}^{QJ}$ between the predicted quality score $q \in [0,1]$ and the ground-truth quality score $\mathbf{Q} \in [0,1]$.
\textit{The ground-truth quality score $\mathbf{Q}$ is obtained by computing the Intersection-over-Union (IoU) between the uncorrected mask $\tilde{\mathbf{M}}$ and the ground-truth mask $\mathbf{M}$, serving as a soft target in the Binary Cross-Entropy loss $\mathcal{L}_{bce}^{QJ}$.}
The mask loss \(\mathcal{L}_{mask}\) is also calculated between the predicted mask \( \hat{\mathbf{M}}\) and the ground-truth mask \( \mathbf{M}\), and it follows the same formulation as that used during MediRound training.
The overall loss \(\mathcal{L}_{JCM}\) can be formulated as:
\begin{equation}
\label{eq:jcm_loss}
\begin{split}
    &\mathcal{L}_{JCM} = \mathcal{L}_{\textit{bce}}^{QJ} + \mathcal{L}_{mask}, \\
    &\mathcal{L}_{mask} = \lambda_{bce}\mathcal{L}_{bce} + \lambda_{dice}\mathcal{L}_{dice}. \\
\end{split}
\end{equation}

\begin{table}[t]
    \footnotesize
    \setlength{\tabcolsep}{0.2cm}
    \centering
    \caption{Comparison of multi-round entity-level medical reasoning segmentation performance between MediRound and existing relevant methods. 
    \textit{``hard case''} denotes a set of cases that are more challenging for models compared with \textit{``regular case''}.
    SegLLM is fine-tuned on MR-MedSeg, whereas the other combined baseline methods are evaluated in a zero-shot manner.
    \(\mathcal{T}\) denotes whether the method is an independent model (a single model or with an extension).
    This table reports the overall scores across all rounds in MR-MedSeg dataset.} 
    \label{table:mt_seg_1}   

\resizebox{\textwidth}{!}{%
        \begin{tabular}{ l | c c c | c c c |c c c | c  }
            \toprule[1.2pt]
            
            \multirow{3}*{\textbf{Methods}} & \multicolumn{3}{c|}{\textbf{val}} & \multicolumn{6}{c|}{\textbf{test}}  & \multirow{3}*{\(\mathcal{T}\)}     \\

            \specialrule{0em}{0pt}{1pt}
            \cline{2-10}
            \specialrule{0em}{0pt}{1pt}

            ~ & \multicolumn{3}{c|}{\textbf{overall}} & \multicolumn{3}{c|}{\textbf{regular case}} & \multicolumn{3}{c|}{\textbf{hard case}} & ~ \\

            \specialrule{0em}{0pt}{1pt}
            \cline{2-10}
            \specialrule{0em}{0pt}{1pt}
            
            ~ & \textbf{Dice} & \textbf{gIoU} & \textbf{cIoU} & \textbf{Dice} & \textbf{gIoU} & \textbf{cIoU} & \textbf{Dice} & \textbf{gIoU} & \textbf{cIoU} & ~  \\ 
            
            \specialrule{0em}{0pt}{1pt}
            \hline
            \hline
            \specialrule{0em}{0pt}{1pt}

            Human-Thinking + BiomedParse \cite{biomed-parse} & 21.1 & 17.0 & 12.1 &28.6  &22.1  &19.0  &11.3  &10.2  &8.4   & \multirow{4}*{\XSolidBrush}\\

            Human-Thinking + MedPLIB \cite{huang2025towards_MedPLIB} &33.7  &25.3  &24.5  &35.0  &24.8  &24.7  &32.0  &25.2  &23.4  & ~ \\
            
            Human-Thinking + IMIS-Net \cite{imis_model} &30.5  &25.9  &32.2    &33.7  &30.0 &42.6  &26.4  &21.0 &25.3  & ~ \\    %

            Human-Thinking + MediSee \cite{tong2025medisee} &43.0  &33.5  &35.4  &45.6  &34.5  &38.6  &45.1 &37.1  &39.7  & ~ \\
            
            \specialrule{0em}{0pt}{1pt}
            \hline
            \specialrule{0em}{0pt}{1pt}
            
            GPT-4o \cite{gpt-4} + IMIS-Net \cite{imis_model} &19.3  &15.8 &19.6  &23.8  &18.7  &20.8  &17.4  &15.2  &21.2  & \multirow{6}*{\XSolidBrush} \\

            Gemini-2.5-Pro \cite{team2023gemini} + IMIS-Net \cite{imis_model} &26.9   &22.6  &28.0 &33.6  &29.5  &35.2  &21.9  &17.0   &19.2  & ~ \\
            
            Qwen3-VL \cite{bai2023qwen} + IMIS-Net \cite{imis_model} &26.6    &22.8  &26.0  &25.1  &22.3  &26.2  &23.4  &18.3  &20.1   & ~ \\

            GPT-4o \cite{gpt-4} + MediSee \cite{tong2025medisee} &36.8  &28.7  &29.4  &45.2  &34.2  &38.0  &33.4  &26.8  &28.9  & ~  \\

            Gemini-2.5-Pro \cite{team2023gemini} + MediSee \cite{tong2025medisee}   &33.5  &26.1  &27.6 &37.9  &31.4  &35.1  &38.6  &28.6  &30.8     & ~ \\
            
            Qwen3-VL \cite{bai2023qwen} + MediSee \cite{tong2025medisee} &42.7  &34.0  &34.5  &45.2  &34.1  &38.0  &37.8  &31.4  &32.8  & ~ \\
            
            \specialrule{0em}{0pt}{1pt}
            \hline
            \specialrule{0em}{0pt}{1pt}
            
            SegLLM-7B \cite{multi-round-seg} & 36.3 & 27.5 & 36.8 & 42.1 & 32.9 & 42.4 & 30.4  & 22.2 & 31.0 & \multirow{2}*{\Checkmark} \\
            
            SegLLM-13B  \cite{multi-round-seg} & 38.9 & 31.2 & 39.1 & 45.3 & 33.4 & 44.0 & 33.0 & 24.6  & 33.9 & ~ \\

            \specialrule{0em}{0pt}{1pt}
            \hline
            \specialrule{0em}{0pt}{1pt}
            
            MediRound + READ \cite{lisa-READ} &53.9  &45.1  &54.6  &60.3  &51.4  &57.6  &47.2  &36.8  &49.1  & ~   \\

            \rowcolor{aliceblue!60}
            MediRound &55.8  &46.5  &55.8  &61.0  &52.1  &58.9  &48.1  &38.2  &50.2 &\Checkmark\\

            \rowcolor{aliceblue!60}
            MediRound + JCM & \textbf{58.4} & \textbf{49.0} & \textbf{58.9} & \textbf{63.3} & \textbf{54.5} & \textbf{60.2} & \textbf{50.3} & \textbf{40.5} & \textbf{53.4} &  \\
            \bottomrule[1.2pt]      
        \end{tabular}
}

\end{table}

\begin{table*}[t]
    \footnotesize
    \setlength{\tabcolsep}{0.1cm}
    \centering
    \caption{
    Multi-round entity-level medical reasoning segmentation performance of MediRound and some relevant methods, reported for each round. The table presents the cIoU scores from rounds 2--8 across all samples in the MR-MedSeg validation set.
    } 
    \label{table:mt_seg_2}   
\resizebox{\textwidth}{!}{%
        \begin{tabular}{ l | c c c  c c c c     }
            \toprule[1.2pt]
            
            \multirow{2}*{\textbf{Methods}} & \multicolumn{7}{c}{\textbf{Conversation Rounds}}      \\ 
            
            \specialrule{0em}{0pt}{1pt}
            \cline{2-8}
            \specialrule{0em}{0pt}{1pt}
            
            ~ & \textbf{Round 2} & \textbf{Round 3} & \textbf{Round 4} & \textbf{Round 5} & \textbf{Round 6} & \textbf{Round 7} & \textbf{Round 8}     \\ 
            
            \specialrule{0em}{0pt}{1pt}
            \hline
            \hline
            \specialrule{0em}{0pt}{1pt}

            Human-Thinking + BiomedParse \cite{biomed-parse} &13.3  &17.9 & 13.9 &15.9  & 11.5  & 9.4  &7.8   \\

            Human-Thinking + MedPLIB \cite{huang2025towards_MedPLIB}  &19.7  &21.0  &21.3  &19.7  &20.5  &25.0  &34.4     \\
            
            Human-Thinking + IMIS-Net \cite{imis_model} &34.0  &27.1  &30.1  &33.4  &37.5  &37.7  &32.8    \\    %

            Human-Thinking + MediSee \cite{tong2025medisee}  &43.6  &37.0  &34.4  &31.9  &29.7  &37.0  &25.1    \\
            
            \specialrule{0em}{0pt}{1pt}
            \hline
            \specialrule{0em}{0pt}{1pt}
            
            GPT-4o \cite{gpt-4} + IMIS-Net \cite{imis_model}  &25.1  &20.4  &18.1  &17.0  &11.6  &12.5  &18.9  \\

            Gemini-2.5-Pro \cite{team2023gemini} + IMIS-Net \cite{imis_model}  &27.2  &28.6  &29.4  &26.2  &22.6  &20.1  &30.5    \\
            
            Qwen3-VL \cite{bai2023qwen} + IMIS-Net \cite{imis_model}  &25.6  &24.1  &20.3  &21.8  &15.2  &14.8  &27.6    \\

            GPT-4o \cite{gpt-4} + MediSee \cite{tong2025medisee}  &37.7  &33.4  &27.3  &25.5  &14.9  &21.5  &20.9  \\

            Gemini-2.5-Pro \cite{team2023gemini} + MediSee \cite{tong2025medisee}  &30.8  &30.4  &29.5  &26.7  &31.5  &21.5  &33.3    \\
            
            Qwen3-VL \cite{bai2023qwen} + MediSee \cite{tong2025medisee}  &37.7  &33.9  &30.1  &25.4  &20.2  &34.7  &29.9    \\
            
            \specialrule{0em}{0pt}{1pt}
            \hline
            \specialrule{0em}{0pt}{1pt}
            
           SegLLM-7B \cite{multi-round-seg}  &35.2  &39.5  &27.2  &30.1  &34.7  &37.4  &34.1   \\
            
            SegLLM-13B \cite{multi-round-seg}  &36.4  &42.1  &32.5  &33.2  &35.3  &39.6  &36.0   \\

            \specialrule{0em}{0pt}{1pt}
            \hline
            \specialrule{0em}{0pt}{1pt}
            
            MediRound + READ \cite{lisa-READ}  &49.1  &58.3  &55.0  &56.2  &62.4  &63.3  &44.5  \\

            \rowcolor{aliceblue!60}
            MediRound  &50.2  &60.8  &55.9  &58.8  &63.7  &64.7  &46.1  \\

            \rowcolor{aliceblue!60}
            MediRound + JCM & \textbf{52.6}  & \textbf{63.6}  & \textbf{59.7}  & \textbf{64.6}  & \textbf{69.5}  & \textbf{66.3}  & \textbf{54.8}     \\
            \bottomrule[1.2pt]      
        \end{tabular}
    }

\end{table*}

\section{Experiments}

\subsection{Experimental Setting}
\noindent \myparagraph{Network Architecture.}
For \textbf{MediRound}, as described in Sec.~\ref{sec:MediRound}, we employ LLaVA-Med-v1.5-Mistral-7B~\cite{llava-med} as the MLLM, and MedSAM~\cite{medsam_model} as the vision backbone. 
The $\mathcal{G}_b^{enc}$ is implemented as a linear projection from 4 dimensions to 4096.
For \textbf{JCM}, the projections in the Quality Judgment Module and Correction Module are implemented as three-layer MLPs with dimensions [256, 512, 512, 1] and [256, 512, 512, 256], respectively.

\noindent \myparagraph{Implementation Details.}
\textbf{MediRound} is trained on the MR-MedSeg using four NVIDIA RTX A6000 GPUs (48 GB memory each) for about 1.5 days.
We use a batch size of 15 on each device.
To enhance training efficiency, we adopt the DeepSpeed \cite{deepspeed} engine. 
We optimize the model parameters with AdamW \cite{adamw} using a learning rate of 0.0003.
The learning rate is adjusted using the WarmupDecayLR scheduler, which includes a 100-iteration warm-up stage.
The LoRA rank is uniformly set to 8. 
For training \textbf{JCM}, we keep the same hyperparameter settings, and it can be completed in just half a day.
All training is conducted on the MR-MedSeg training set, employing a teacher forcing strategy \cite{teacher_forcing}.
During training, the weights for the Dice loss \(\lambda_{dice}\) and BCE loss \(\lambda_{bce}\) are set to 0.5 and 2.0.
All multi-round segmentation evaluations adopt an autoregressive free-running strategy, where subsequent rounds can only rely on the model’s own mask outputs from previous rounds.

\subsection{Multi-Round Experimental Results}
The results of multi-round medical reasoning segmentation are summarized in Tables \ref{table:mt_seg_1} and \ref{table:mt_seg_2}.
Table \ref{table:mt_seg_1} reports the overall performance across all rounds, while Table \ref{table:mt_seg_2} provides detailed scores for each individual round of interaction.
Since existing text-prompt-based medical segmentation methods lack the capability for multi-round entity-level reasoning segmentation, we first incorporate human guidance to enable a broader and more equitable comparison. 
Specifically, humans with medical expertise quickly review the multi-round questions and historical outputs, infer the target at each round, then write a textual description as the prompt to the segmentation models, allowing these models to perform the task.
As shown in Tables \ref{table:mt_seg_1} and \ref{table:mt_seg_2}, human assistance can substantially improve the model’s ability to perform multi-round segmentation. However, the overall performance remains suboptimal. This is primarily because, even with a certain level of medical knowledge, humans may still make errors when dealing with complex multi-round medical reasoning tasks, and their coordination with the model tends to be relatively rigid.
In contrast, the proposed MediRound can effectively comprehend the current query while simultaneously integrating historical information and mask results from referred rounds. 
By achieving this capability, it yields an average improvement of approximately 15\% across various evaluation metrics compared to other methods.
Notably, despite its lightweight design, the proposed JCM remarkably improves the model’s overall performance, particularly as the number of interaction rounds increases (refer to Table \ref{table:mt_seg_2}).
This improvement is primarily attributed to the fact that JCM refines suboptimal mask outputs from earlier rounds before they are referred in subsequent ones, thereby effectively mitigating the error accumulation inherent in the chain-like pipeline of multi-round segmentation.

\begin{wraptable}{r}{0.5\textwidth}
\centering
\caption{Vanilla medical referring segmentation comparison results.}
\label{table:traditional_seg}
\footnotesize
\tabcolsep=0.12cm
\resizebox{0.5\textwidth}{!}{%
\begin{tabular}{ l | c c c }
\toprule[1.2pt]
\multirow{2}*{\textbf{Methods}} & \multicolumn{3}{c}{\textbf{Metrics}}\\
\cline{2-4}
~ & \textbf{Dice} & \textbf{gIoU} & \textbf{cIoU} \\
\hline\hline
\multicolumn{4}{l}{\emph{\textbf{Medical Interaction Seg. (bbox input)}}} \\
\largemodel MedSAM \cite{medsam_model} & \largemodel 77.7 & \largemodel 67.1 & \largemodel 81.0 \\
\largemodel IMIS-Net \cite{imis_model} & \largemodel 77.0 & \largemodel 67.5 & \largemodel 87.2 \\
\hline
\multicolumn{4}{l}{\emph{\textbf{Medical Referring Seg.}}} \\
Grounding DINO \cite{grounding-dino} + SAM-Med2D \cite{sam-med-2d-model} & 18.4 & 12.5 & 10.9 \\
Grounding DINO \cite{grounding-dino} + MedSAM \cite{medsam_model} & 18.5 & 11.0 & 13.5 \\
BiomedParse \cite{biomed-parse} & 24.9 & 20.1 & 14.5 \\
MedPLIB \cite{huang2025towards_MedPLIB} & 40.6 & 39.1  & 45.2 \\
IMIS-Net \cite{imis_model} & 41.3 & 35.5 & 65.8 \\
MediSee \cite{tong2025medisee} & 61.2 & \textbf{50.4} & \textbf{70.8} \\
MediRound + READ \cite{lisa-READ} & 60.7 & 48.0 & 65.2 \\
\rowcolor{aliceblue!60}
MediRound & \textbf{62.1} & 48.3 & 66.3 \\
\bottomrule[1.2pt]
\end{tabular}%
}
\end{wraptable}

We further construct a hybrid paradigm for comparison, which integrates an MLLM with a medical segmentation model.
Specifically, we leverage the strong text–image reasoning capabilities of MLLMs \cite{bai2023qwen,gpt-4,team2023gemini} to interpret complex multi-round queries, and subsequently feed their generated reasoning outputs into a medical segmentation model to produce visual predictions.
Notably, all comparison methods are also allowed to access the historical mask outputs to ensure a fair comparison.
As shown in Tables \ref{table:mt_seg_1} and \ref{table:mt_seg_2}, despite being supported by powerful MLLMs, these combined approaches still underperform compared with MediRound.
We attribute this advantage to our model’s end-to-end training on extensive medical image–text pairs, which enables more coherent dynamic alignment between multimodal features than the two-stage combined paradigms.
In addition, we include SegLLM \cite{multi-round-seg}—the exploratory work on multi-round reasoning segmentation in natural images—in our comparison.
The results indicate that while SegLLM performs well in natural images, it struggles when generalized to medical scenarios.

\begin{table}
    \centering
    \begin{minipage}[t]{0.55\linewidth}
        \centering
        \scriptsize
        \setlength{\tabcolsep}{0.1cm}
        \caption{Ablation study on the choices of vision and MLLM backbones. Experiments are conducted on the MR-MedSeg val set.}
        \label{table:mian_component}
        \begin{tabular}{l | c c c }
            \toprule[0.9pt]
            \textbf{Settings}  & \textbf{Dice}& \textbf{gIoU} & \textbf{cIoU}  \\
            \midrule
            \midrule
            \multicolumn{4}{l}{\emph{{\textbf{Seg. Head Backbone}}}} \\
            SAM & 46.7 & 33.1   & 37.2 \\
            SAM-Med2D  & 53.9 & 45.3   & 55.5   \\
            MedSAM (LoRA) & 54.6 & 44.9   & 54.4   \\
            \rowcolor{aliceblue!60}
            MedSAM  & 55.8 & 46.5   & 55.8 \\
            \specialrule{0em}{0pt}{1pt}
            \hline
            \specialrule{0em}{0pt}{1pt}
            \multicolumn{4}{l}{\emph{{\textbf{MLLM Backbone}}}} \\
            LLaVA-v1.5 (LoRA) & 53.4 & 42.1   & 53.3 \\
            LLaVA-Med (Frozen) & 48.0 & 39.2   & 49.6 \\
            \rowcolor{aliceblue!60}
            LLaVA-Med (LoRA) & 55.8 & 46.5   & 55.8 \\
            \bottomrule[0.9pt]
        \end{tabular}
        
    \end{minipage}
    \hfill
    \begin{minipage}[t]{0.43\linewidth}
        \centering
        \footnotesize
        \setlength{\tabcolsep}{0.25cm}
        \caption{Ablation study on the effects of different strategies for extracting reference-mask information. 
        The experimental results are reported on the val set of MR-MedSeg.
        \(\mathcal{C}\): Cropped image input; \(\mathcal{B}\): Bounding box input.}
        \label{appendix_table:input_ref}

        \resizebox{\linewidth}{!}{
        \begin{tabular}{c c | c c c }
            \toprule[1.2pt]
            \(\mathcal{C}\) & \(\mathcal{B}\)   & \textbf{Dice} & \textbf{gIoU} & \textbf{cIoU} \\
            \midrule
            \midrule
            \XSolidBrush & \XSolidBrush & 41.9 & 32.5  & 38.4  \\
            \Checkmark & \XSolidBrush & 55.4 & 45.8 & 56.0   \\
            \rowcolor{aliceblue!60}
            \Checkmark & \Checkmark  & 55.8 & 46.5  & 55.8  \\
            \bottomrule[1.2pt]
        \end{tabular}
        }
    \end{minipage}

\end{table}

\subsection{Vanilla Medical Referring Segmentation}
To demonstrate MediRound's effectiveness in conventional single-round referring medical segmentation, we evaluate it against several prior related methods on the SA-Med2D-20M \cite{sam_med_2d_20m} benchmark.
As summarized in Table \ref{table:traditional_seg}, our model achieves strong performance, remaining highly competitive in traditional medical image segmentation.
For context, we also report the segmentation results of IMIS-Net \cite{imis_model} and MedSAM \cite{medsam_model} using conventional bounding box guidance.

\subsection{Ablation Study}
\begin{figure}[t]
    \centering
    \includegraphics[width=0.8\linewidth]{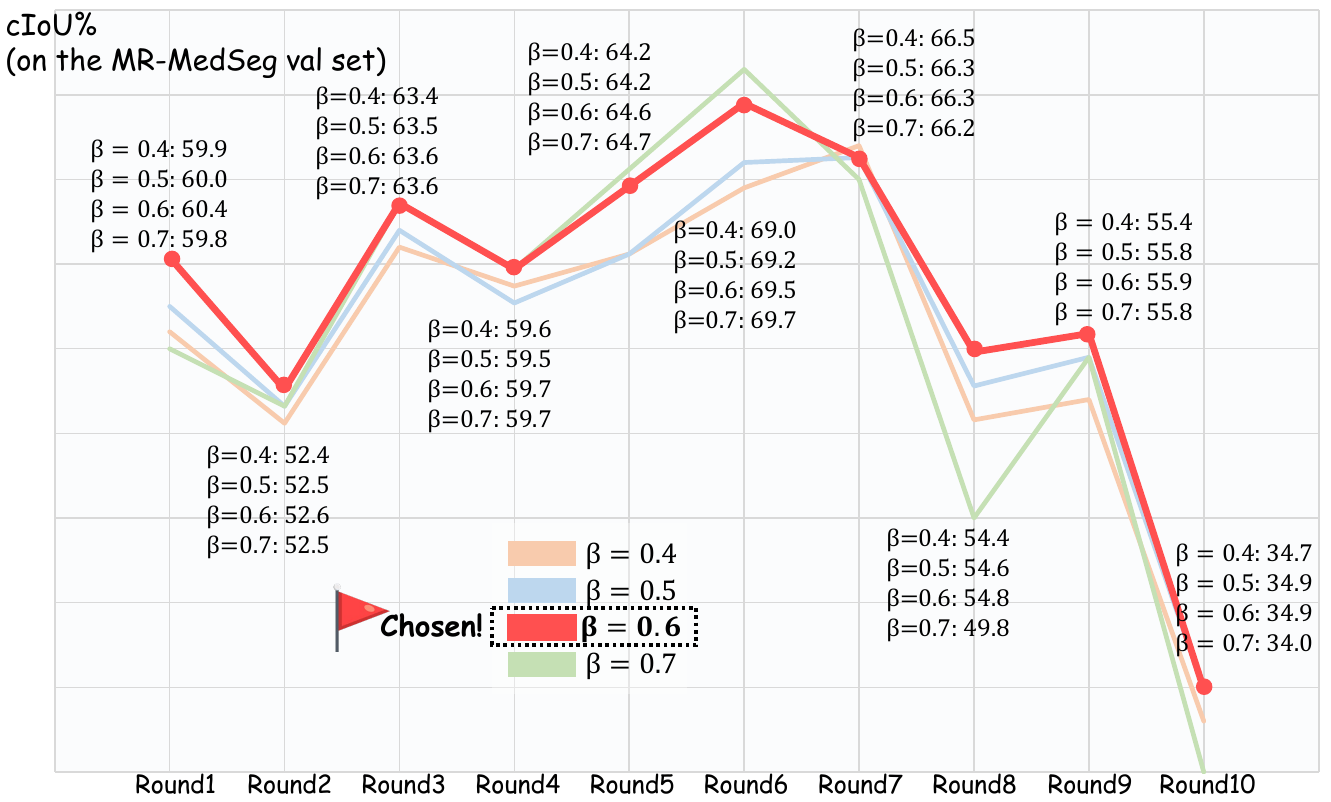}
    \caption{Ablation study on the threshold \( \beta \) value in JCM.}
    \label{fig:jcm_ab}
\end{figure}

\noindent \myparagraph{Choices of \(\boldsymbol{\beta}\) in JCM.} 
We investigate the impact of the threshold \( \beta \) in the Quality Judgment Module of JCM.
As shown in Figure~\ref{fig:jcm_ab}, the collaboration between JCM and MediRound achieves the best performance when $\boldsymbol{\beta = 0.6}$. This is likely because a smaller \( \beta \) causes many suboptimal results to lose the opportunity for refinement, while an excessively large \( \beta \) may instead disturb some high-quality predictions.
\textit{For the definition and role of $\beta$, see Algorithm \ref{alg:multi_round_seg_jcm}.}

\noindent \myparagraph{Choices of the Main Components.} 
Table~\ref{table:mian_component} summarizes the effects of different foundational MLLMs and segmentation head backbones. The results indicate that \textbf{LLaVA-Med} achieves superior performance owing to its strong multimodal alignment capability in medical contexts, while the \textbf{MedSAM} performs best as the segmentation head, benefiting from large-scale medical image pretraining.

\noindent \myparagraph{Choices of Reference Information Strategies.} 
We further examine the impact of different strategies for extracting information from the reference mask. Table \ref{appendix_table:input_ref} indicates that using both the cropped image and the bounding box as reference-mask information yields the best performance.

\subsection{Qualitative Analysis}
\begin{figure*}[t]
    \centering
    \includegraphics[width=1\linewidth]{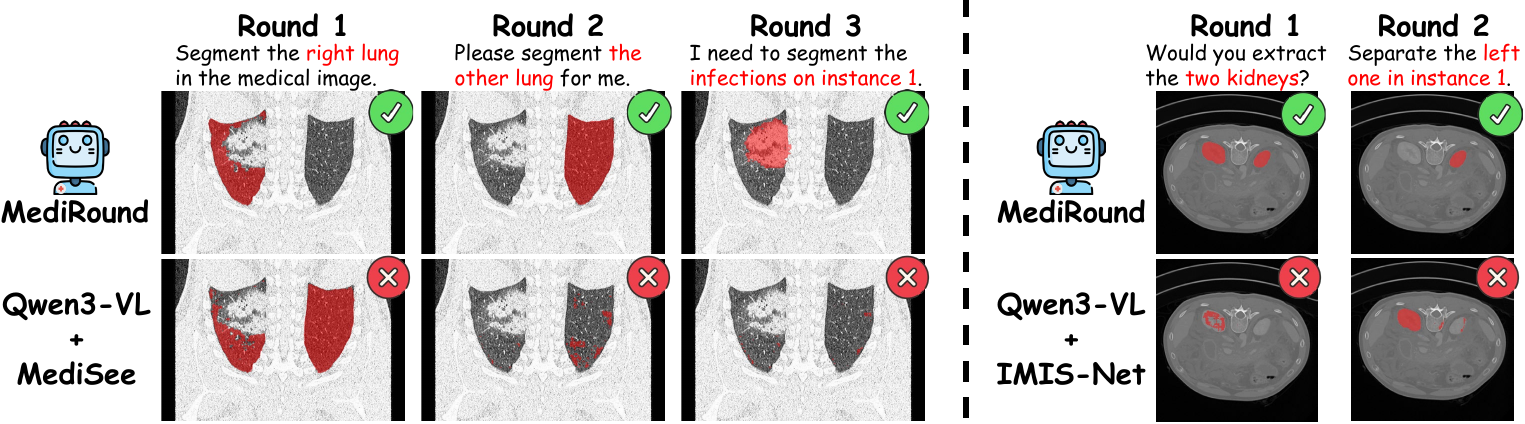}
    \caption{
    Qualitative comparison with different methods.
    }
    \label{fig:visual_perform_exp}
\end{figure*}

We present qualitative comparisons in Figure~\ref{fig:visual_perform_exp}. Existing methods largely fail in multi-round reasoning for medical segmentation, whereas MediRound can understand the current query and dialogue history for accurate segmentation.

\subsection{Human Evaluation}
\begin{figure*}
    \centering
    \includegraphics[width=0.9\linewidth]{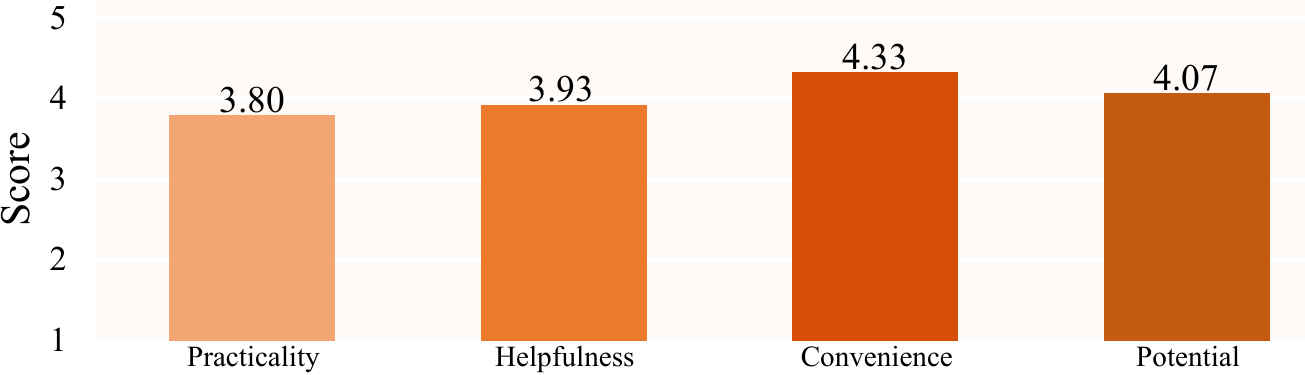}
    \caption{Human evaluation on MEMR-Seg and MediRound (1–5 scores).
    This experiment is designed to demonstrate the potential value of our work in medical education scenarios.
    }
    \label{fig:human_eval}
\end{figure*}

We recruited 15 medical students to rate our proposed multi-round entity-level medical reasoning segmentation task (MEMR-Seg) and its baseline, MediRound, on four criteria using a 5-point scale (1–5).
The results in Figure \ref{fig:human_eval} demonstrate the task and model’s meaningfulness and practical value in medical education.

\section{Conclusion}
This work introduces MEMR-Seg, a new task that advances medical image segmentation toward interactive, multi-round reasoning. By constructing the large-scale MR-MedSeg dataset and developing the MediRound baseline with a helpful Judgment \& Correction Mechanism, this work demonstrates the feasibility and effectiveness of medical multi-round entity-level reasoning, providing insights for future research on interactive medical segmentation.

\bibliographystyle{splncs04}
\bibliography{main}

\clearpage

\end{document}